%% file: main.tex
\pdfoutput=1
\documentclass[11pt]{article}

\usepackage[review]{ACL2023}

\usepackage{times}
\usepackage{latexsym}

\usepackage[T1]{fontenc}
\usepackage[utf8]{inputenc}

\usepackage{microtype}

\usepackage{inconsolata}

\usepackage{booktabs}
\usepackage{graphicx}
\usepackage{subcaption}
\usepackage{amsmath}
\usepackage{geometry}
\usepackage[linguistics]{forest}
\usepackage{natbib}
\usepackage{hyperref}
\usepackage{afterpage}
\usepackage{adjustbox}
\usepackage{amssymb}

\title{Exploring Multi-Modality Dynamics: Insights and Challenges in Multimodal Fusion for Biomedical Tasks }
\author{Laura Wenderoth \\
  Newnham College \\
  University of Cambridge\\
  {\tt lw754 (at) cam.ac.uk} \\}

\begin{document}
\maketitle
\begin{abstract}
\input{sections/01_Abstract}
\end{abstract}

\input{sections/02_Introduction}

\input{sections/02_MM_dynamics}

\input{sections/03_Methods}

\input{sections/04_Results}

\input{sections/05_Discussion}

\bibliography{project}
\bibliographystyle{acl_natbib}

\appendix
\onecolumn
%\newpage
\section{Appendix}
\label{sec:appendix}

\input{sections/09_Appendix}

\end{document}

%% file: sections/01_Abstract.tex
This paper investigates the MM dynamics approach proposed by \citet{Han_Yang_Huang_Zhang_Yao_2022} for multi-modal fusion in biomedical classification tasks. The MM dynamics algorithm integrates feature-level and modality-level informativeness to dynamically fuse modalities for improved classification performance. However, our analysis reveals several limitations and challenges in replicating and extending the results of MM dynamics. We found that feature informativeness improves performance and explainability, while modality informativeness does not provide significant advantages and can lead to performance degradation. Based on these results, we have extended feature informativeness to image data, resulting in the development of Image MM dynamics. Although this approach showed promising qualitative results, it did not outperform baseline methods quantitatively.\\

%% file: sections/02_Introduction.tex
\section{Introduction}
%The importance of processing multi-modal data in medicine is growing due to the inherently multi-modal nature of the field \cite{Acosta_Falcone_Rajpurkar_Topol_2022, Shaik_Tao_Li_Xie_Velásquez_2024}.  Key modalities include tabular genetic data, histopathological images, MRI and CT scans, as well as EEG and ECG time-series data  \cite{Azam_Ryabchykov_Bocklitz_2022}. The aim is to create a machine learning application that is both precise and explainable, incorporating all relevant modalities for a given task.
The field of medicine is inherently multi-modal, which is why processing multi-modal data has become increasingly important \cite{Acosta_Falcone_Rajpurkar_Topol_2022, Shaik_Tao_Li_Xie_Velásquez_2024}. This involves working with various forms of data such as tabular genetic data, histopathological images, MRI and CT scans, EEG and ECG time-series data \cite{Azam_Ryabchykov_Bocklitz_2022}. The ultimate goal is to develop machine learning applications that are both precise and explainable, incorporating all relevant data modalities for a given task.
 
However, the fusion of multiple modalities poses several challenges.
One of the significant challenges is determining when to fuse the modalities \cite{Azam_Ryabchykov_Bocklitz_2022}. There are different fusion methods, including early fusion, late fusion, and intermediate fusion each with its own advantages and disadvantages  \cite{Stahlschmidt_Ulfenborg_Synnergren_2022}. Early fusion involves concatenating features before feeding them into the model and is typically only viable when the data share  similar dimensions. Late fusion processes modalities in modality-specific encoders and concatenates the latent space representations before training a shared decoder. It has the advantage of utilising pre-existing uni-modal models, whose effectiveness has already been assessed. However, it is limited in its ability to learn interactions between modalities \cite{Pawłowski_Wróblewska_Sysko-Romańczuk_2023}. In contrast, intermediate fusion aims to strike a balance by enabling the learning of uni-modal representations while also capturing interactions between modalities.

The second challenge is deciding which method is used to perform the fusion. Early fusion often uses a simple concatenation without weighting, while late fusion requires careful design decisions. 
\citet{Pawłowski_Wróblewska_Sysko-Romańczuk_2023} emphasise the importance of selecting the appropriate fusion technique to create a multimodal representation that achieves optimal modelling performance. Various methods exist for fusing uni-modal representations, such as direct concatenation, pre-weighting, or selecting the most significant modality per data point \cite{Stahlschmidt_Ulfenborg_Synnergren_2022}.

Previous methods have utilised static fusion strategies \cite{Dimitrovski_Kocev_Kitanovski_Loskovska_Džeroski_2015,Stahlschmidt_Ulfenborg_Synnergren_2022,Pawłowski_Wróblewska_Sysko-Romańczuk_2023}, which may not be suitable for biomedical applications. It has been argued that static fusion strategies are insufficient for biomedical applications due to the varying importance of different modalities for each patient. For example, when analysing the survival of cancer patients, the importance of modalities like genetic data and histopathological images may vary based on the patient's subtype or condition. Therefore, it is desirable to fuse these modalities dynamically. Additionally, maintaining a high level of explainability is crucial in biomedical applications.  A desirable approach would be inherently explainable and incorporates various modalities while also making dynamic decisions about the importance of each modality for the individual patient.

Multimodal Dynamics \cite{Han_Yang_Huang_Zhang_Yao_2022} combines these requirements. It is a late fusion technique that identifies the most informative modality for each patient, achieving state-of-the-art results on four genetic multimodal data sets. This approach is dynamic and explainable because it integrates built-in explainability to assess the informativeness of each modality and its constituent features. The main hypotheses tested in the original paper are the following:

\begin{itemize}
    \item [(a)] Dynamical fusion outperforms state-of-the-art results and is explainable at the same time.
    \item [(b)] The informativeness of each modality for the downstream task varies among samples but can be predicted.
    \item  [(c)] Reducing the number of features improves classification by reducing noise.
\end{itemize}

The first hypothesis \textit{(a)} was clearly confirmed by \citet{Han_Yang_Huang_Zhang_Yao_2022}. They have demonstrated the effectiveness of their pipeline on four distinct data sets, achieving state-of-the-art results and surpassing them on three of the data sets. In addition to the quantitative analysis, they also carried out a qualitative analysis to show the explainability of their approach. The analysis revealed KLK8 as the most important feature. KLK8 is downregulated in breast cancer and is an independent indicator of prognosis, according to the literature. This proves that the features identified as important are indeed important for cancer progression, according to the biomedical literature. 
The second hypothesis \textit{(b)} was predominantly confirmed. The authors demonstrated that their model weighted the modalities differently depending on the sample. It was found that, on average, one modality held more significance than the other two. However, an ablation study demonstrated that dynamical fusion improved the overall result. This clearly indicates that the model weighted the modalities with varying degrees of informativeness for different samples. The authors did not further investigate the quality of modality informativeness prediction. They solely argued that their result outperformed the state-of-the-art, implying the validity of their prediction.
In addition, the authors predominantly confirmed the third hypothesis \textit{(c)}. The ablation study demonstrated that feature weighting improves classification performance, which helps filter important features while reducing noise.

However, it is important to note that MM dynamics has limitations as it was only tested on tabular genetic data. This is a significant limitation as biomedical data often includes images and time-series data. Furthermore, the original paper lacked some necessary ablation studies, which are explained in detail in Section \ref{subsec:limitations}.
This paper reproduces MM dynamics and extends it to images, providing the missing ablation studies and comparisons. The most important contributions can be summarised as follows:

\paragraph{}
\begin{itemize}
    \item[\textbf{i)}]  \textbf{Image MM Dynamics}: MM dynamics is extended from tabular genetic data to images. The main challenge is to adapt the feature informativeness estimation to preserve interpretability. Using U-nets, we introduce a method to calculate feature informativeness in a meaningful way, not at the pixel level but at the patch level.
    \item [\textbf{ii)}]  \textbf{Application of a different data set}: The aim is to test the main hypothesis of the original paper, which suggests that dynamic fusion leads to state-of-the-art results. To accomplish this, we selected a new data set \cite{open-problems-multimodal, matek2021expert} comprising protein and RNA data, as well as histopathological images from single cells extracted from bone marrow. These modalities are used to identify the single cells and their corresponding biological classes.
    \item [\textbf{iii)}]  \textbf{Investigation of the effectiveness of feature and modality informativeness}:
    Through extensive ablation studies and additional experiments, the usefulness of feature and modality informativeness on classification performance will be investigated. This involves investigating how accurately the informativeness of each modality can be estimated to fully address hypothesis \textit{(b)}. Furthermore, it conducts additional investigations into the informativeness of features to better establish whether hypothesis \textit{(c)} is valid.
\end{itemize}

Section \ref{section:mmdynamics} describes in detail the relevant methodology of the original paper, providing a comprehensive understanding of the approach. Section \ref{section:methods} introduces the data set and outlines the methodology for expanding the classification to images. The results are presented in Section \ref{section:results}, which includes an in-depth examination of ablation studies to clarify the findings further. Section \ref{section:discussion} provides an outlook and discussion, offering insights into potential future research directions and discussing the implications of the results obtained.

Additionally, ensuring the reproducibility and transparency of our research is of utmost importance to us. Therefore, the implementation and data can be accessed at \href{https://github.com/LauraWenderoth/MMdynamics-evolved}{\texttt{LauraWenderoth/MMdynamics-evolved}}.

%% file: sections/02_MM_dynamics.tex
\section{Multimodel dynamics} 
\label{section:mmdynamics}

The Multimodal Dynamics algorithm is a multimodal classification approach that employs a dynamical fusion method that exploits the informativeness of features and modalities. It is based on the work of \citet{Tonge_Caragea_2019}, who identified reliable modalities for text-image pairs. The algorithm uses an additional network developed by \citet{Panda_Chen_Fan_Sun_Saenko_Oliva_Feris_2021} to identify important features (time frames) for each modality (video, audio) in every sample. The Multimodal Dynamics algorithm integrates these techniques and applies them to the novel modality of genetic tabular data. 

\begin{figure*}
    \centering
    \includegraphics[width=\linewidth]{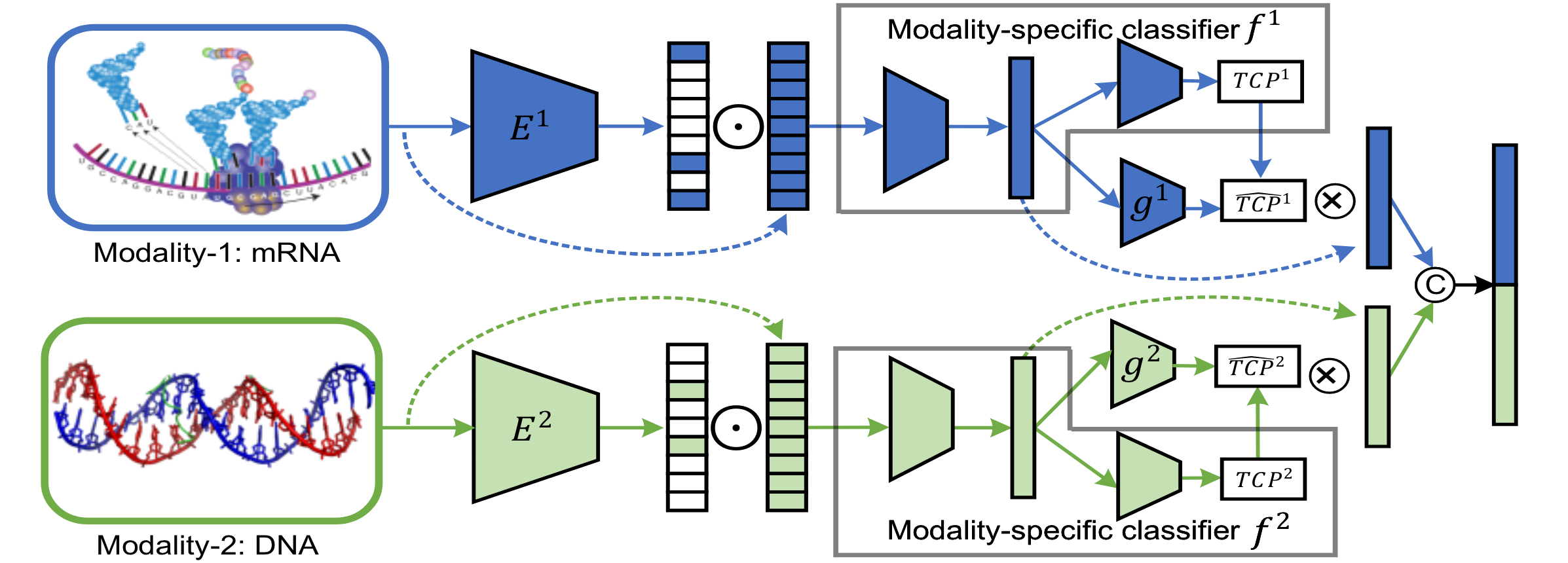}
    \caption{Schematic overview of the MM dynamics algorithm. The schematic structure of the MM dynamics algorithm consists of three main steps. Firstly, the feature informativeness is computed using the encoder \(E^m\), where \(m\) ranges from 1 to \(M\), representing the number of modalities. Subsequently, the input is multiplied by the feature informativeness vector and serves as input for the uni-modal classifier \(f^m\). The second step involves calculating the modality informativeness, also known as the true class probability ($TCP$), using the regression networks \(g^m\). The latent representation of \(g^m\) is then weighted based on the estimated $\widehat{TCP}$. In the final step, the dynamically weighted representations are concatenated, and the ultimate classification is performed using another classifier not illustrated in the figure.   Reprinted from \citet{Han_Yang_Huang_Zhang_Yao_2022}.}
    \label{fig:overview_mmdynamics_algo}
\end{figure*}

\subsection{Algorithm}

The algorithm consists of three main components. A schematic overview of the algorithm is shown in Figure~\ref{fig:overview_mmdynamics_algo}. The first component is a modality-specific encoder with integrated feature informativeness, the second is a weighting of the modality-specific latent representation using the modality informativeness, and finally, the third is the actual classification by concatenating the weighted representations and feeding them into a classifier. Likewise, the loss function consists of three parts and is defined by:
\begin{equation}
    \mathcal{L}=\sum_{i=1}^N\left(\lambda_1 \mathcal{L}^{\ell_1}+\lambda_2 \mathcal{L}^{\text {conf}}+\lambda_3 \mathcal{L}^f\right)
\end{equation}
where $N$ is the size of the data set, $\mathcal{L}^{\ell_1}$ is the modality-specific feature informativeness loss, $\mathcal{L}^{\text {conf}}$ the modality informativeness loss,  $\mathcal{L}^f$ the final classification loss and  $\lambda_1,\lambda_2,\lambda_3$ are  the corresponding weights. The individual loss functions are explained in more detail in the following.

\paragraph{Feature-level informativeness} The concept of feature-level informativeness asserts that every feature within a data set may hold varying degrees of significance. Insignificant features have the potential to add unwanted noise, thereby obstructing the extraction of valuable insights. With this in mind, the authors propose that filtering out irrelevant features prior to training uni-modal classifiers is a more effective strategy and can lead to improved outcomes. A one-layer fully connected neural network $E^m:\mathbf{x}^m \rightarrow \mathbf{w}^m$ is used to accomplish this task, where $\mathbf{w}^m \in \mathbb{R}^{d_m}$. The network takes the features of a modality $\mathbf{x}^m$ as input and outputs a number $\mathbf{w}^m_i$ for each feature $\mathbf{x}^m_i$. The larger the number, the more important the feature is considered. To reduce the number of features, an $L_1$ norm is used as the loss function:

\begin{equation}
\mathcal{L}^{\ell_1}=\sum_{m=1}^M\left\|\mathbf{w}^m\right\|
\end{equation}

where $M$ is the number of different modalities. The outputs $\mathbf{w}^m$ are multiplied with $\mathbf{x}^m$ to suppress uninformative features: $\widetilde{\mathbf{x}}^m=\mathbf{x}^m \odot \mathbf{w}^m$, where $ \odot$ represents the element-wise multiplication. The result $\widetilde{\mathbf{x}}^m$ is then used as an input to the uni-modal encoder. The outputs $\mathbf{w}^m$  can finally be used to determine which features, in this case gene expressions, were most important for the prediction. This step provides the inherent explainability of the approach.

\paragraph{Modality-level informativeness} The assumption behind modality-level informativeness is that the more confidently a correct prediction is made by a uni-modal classifier, the greater the informativeness of that modality. To achieve this, two steps are required:  training an unimodal classifier, and  training a regression network to predict the confidence level of the unimodal classifier's predictions.
First, this requires the training of $M$ uni-model classifiers $f^m$. The objective $\mathcal{L}^{c l s}$ of the training  is to minimize the relative entropy between the predicted distribution $\mathbf{p}^m(\mathbf{y} | \mathbf{x}^m)$ based on the softmax output and the true labels $\mathbf{y}$:

\begin{equation}
\mathcal{L}^{c l s}=-\sum_{m=1}^M  \mathbf{y } \log \mathbf{p}^m(\mathbf{y} | \mathbf{x}^m),
\end{equation}

To quantify the confidence of the prediction, the true class probability per modality $TCP^m$ is utilised:
\begin{equation}
T C P^m=\mathbf{y} \cdot \mathbf{p}^m\left(\mathbf{y} \mid \mathbf{x}^m\right),
\end{equation}
where $\mathbf{p}^m(\mathbf{y} | \mathbf{x}^m)$ are the softmax outputs of the uni-model classifier. Multiplying this by the ground truth label $\mathbf{y}$ results in the probability for the correct label, known as the $TCP^m$.

During testing, the ground truth labels $\mathbf{y}$ are unavailable. Therefore, the $TCP^m$ cannot be directly computed. To address this, a regression network $g^m$ is used to calculate the estimated TCP:
\begin{equation}
\widehat{T C P}^m=g^m\left(\mathbf{x}^m\right),
\end{equation}
where $\mathbf{x}^m$ are the input features. The regression network is a single-layer feed forward neural network with a single numerical output. This regression network is trained by minimizing the $L_2$ distance between $TCP^m$ and $\widehat{T C P}^m$. Now, combine the two parts to obtain the following loss function for the modality confidence:

\begin{equation}
\mathcal{L}^{\text {conf }}=\sum_{m=1}^M\left(\widehat{T C P}^m-T C P^m\right)^2+\mathcal{L}^{c l s}
\label{eq:confidenceloss}
\end{equation}

\paragraph{Fusion and Classification}
After defining the uni-model classifiers $f^m$, their latent representations are used in the multimodal dynamic fusion process. Therefore, the last fully connected layer is removed from $f^m$ to obtain $f^m_1$, which outputs the latent representation $\mathbf{h}^m=f_1^m\left(\widetilde{\mathbf{x}}^m\right)$. These representations are then weighted by multiplying them by the estimated TCP of each modality. Finally, the modalities are concatenated as follows:

\begin{equation}
\mathbf{h}=\left[\widehat{T C P}^1 \mathbf{h}^1, \cdots, \widehat{T C P}^M \mathbf{h}^M\right]
\end{equation}

where $[\cdot, \cdot]$ is the concatenation operator. Finally, a classifier $f : \mathbf{h} \rightarrow \mathbf{y}$ is trained. This classifier is again a uni-layer feed-forward neural network with the output shape equal to the number of desired classes. The cross-entropy loss function is chosen, resulting in the classification loss $\mathcal{L}^f$:

\begin{equation}
\mathcal{L}^f = -\mathbf{y }\log \mathbf{p}\left(\mathbf{y} \mid \mathbf{x}\right),
\end{equation}

where $\mathbf{y }$ and $ \mathbf{p}\left(\mathbf{y} \mid \mathbf{x}\right)$ are vectors representing the ground truth and predicted probabilities, respectively.

\subsection{Limitations} 
\label{subsec:limitations}
The MM dynamics approach has shown great potential, achieving state-of-the-art results on four tabular genetic data sets.
However, compared to twelve baseline approaches, the late fusion methods were absent despite the paper's focus on dynamic late fusion. 
%Therefore the question about the suitability of late fusion on these data sets compared to early or intermediate fusion methods can not be answered. 
The authors did not compare their approach to static weighted late fusion, which is a significant omission because they could not answer the question of whether a dynamic late fusion is better than a static late fusion. 
%In one of their ablation studies, they only compared against MM dynamics without modality and feature informativeness, which is an unweighted static late fusion. They demonstrated the superiority of dynamic late fusion in this unfair comparison.

In addition, the author's assertion about the relationship between the classification confidence of modalities $(\widehat{T C P})$ and their informativeness lacks empirical support. Their claim of being able to predict classification confidence $(\widehat{T C P})$ remains unsubstantiated due to the absence of supporting evidence or ablation study results.
If classification confidence $(\widehat{T C P})$ is indeed effective, it should also identify and filter out noisy modalities. However, the paper does not address handling missing modalities, which is a crucial aspect to consider. 

Furthermore, the paper lacks many details necessary for reproducibility. Despite the authors publishing the code, it is a pared-down version covering only the MM dynamics model and two of the four data sets, consisting of test and train split. The code does not generate the validation split, and the hyperparameters are arbitrarily set without explanation or elaboration in the paper. The lack of transparency regarding essential hyperparameters, such as the dimension of the latent representation of the uni-modal encoder and the weighting of different parts of the loss function, as well as the absence of a validation data set, raises concerns about potential overfitting on the test set. Additionally, baseline methods were not implemented, making it impossible to validate them.

The primary limitation lies in the restricted applicability of the findings to tabular data alone, without addressing the potential adaptation of the approach to other modalities like images, text, or audio, which are fundamental to biomedical classification tasks.

Addressing these limitations enhances the research's significance. This involves providing empirical evidence to support claims, addressing critical considerations such as missing modalities, ensuring transparency in reporting essential details for reproducibility, including using a validation data set, implementing baseline methods for comparison, and exploring the approach's adaptability to diverse modalities beyond tabular data.

%\colorbox{red}{Allgemeines ziel einer klassification: given features for M modalites and coressepondig labels (formula) ziel ist mapping funktion zu kreieren}

%% file: sections/03_Methods.tex
\section{Methods}
\label{section:methods}

To replicate MM dynamics, we initially focused solely on genetic data. The dataset used is described in Subsection \ref{subsection:data}. Additionaly, we implemented a variety of baselines and provided a rationale for why they were chosen. The evaluation is performed in a patient-specific manner. The dataset is split between most patients for training and validation, while the remaining patients are reserved for testing. This split ensures a fair comparison of methods as overfitting on patients is bypassed. The presentation of the novel extension of the methodology to incorporate images, specifically Image MM dynamics, concludes this section.

\subsection{Data}

\label{subsection:data}
The dataset created by \citet{open-problems-multimodal} consists of multi-omics data such as gene expression profiles (RNA) and protein levels (protein) of single cells from mobilised peripheral CD34+ haematopoietic stem and progenitor cells isolated from three different healthy human donors. These cells are vital to the hematopoietic system and can differentiate in the bone marrow into different blood cells. Each cell is annotated with its corresponding cell type which cover B-lymphocytes progenitor, neutrophil progenitor, monocyte progenitor, and erythrocyte progenitor.

In addition, an independent image dataset was selected, comprising single-cell images from bone marrow \cite{matek2021expert}. From this dataset, images corresponding to the four classes — lymphocytes, neutrophilic granulocytes, monocytes, and erythroblasts — were respectively chosen and are illustrated in Figure \ref{fig:dataset_images}. Each multi-omics data point is associated with an image depicting the same cell type. It is worth noting that the images presented may not necessarily depict the exact cells from which the omics data was extracted.  Nevertheless, this disparity does not present any significant obstacle, given that the primary objective is to classify cell types rather than to match individual cells across various modalities.

\input{Figures/data_set_figure}

\subsection{Baselines}
\input{Figures/main_results_table_rna_protein}
When selecting baseline methods for comparison, we considered early, intermediate, and two late fusion approaches. Early fusion methods, such as k-nearest neighbours (KNN), logistic regression (LR), neural networks (NN), random forests (RF), and support vector machines (SVM), concatenate features from different modalities before inputting them into the classifier. For intermediate fusion, we selected Hybrid Early-fusion Attention Learning Network (HEALNet \cite{Hemker_Simidjievski_Jamnik_2023HEALNet}), which was specifically designed for handling histopathological images and genetic data. For late fusion methods, we chose  Multimodal Co-Attention Transformer (MCAT) by \citet{Chen_Lu_Weng_Chen_Williamson_Manz_Shady_Mahmood_2021mcat}, which is tailored for histopathological images and genetic data, and we also chose a static multimodal fusion with the same uni-modal encoder as MM dynamics.

\paragraph{HEALNet}  by \citet{Hemker_Simidjievski_Jamnik_2023HEALNet} is a multi-modal fusion architecture that aims to overcome the limitations of modality-specific approaches. It achieves this by preserving modality-specific structural information, capturing cross-modal interactions in a shared latent space, and effectively handling missing modalities during training and inference. HEALNet demonstrates state-of-the-art performance in multi-modal survival analysis on histopathological images and genetic data, outperforming both uni-modal and recent multi-modal baselines while maintaining robustness in scenarios with missing modalities and is therefore chosen as the baseline.

\paragraph{MCAT}  introduced by \citet{Chen_Lu_Weng_Chen_Williamson_Manz_Shady_Mahmood_2021mcat} is used to predict survival outcomes in computational pathology. The framework leverages dense co-attention mapping between histopathological images and genomic features to model complex interactions within the tumour microenvironment effectively. Based on Visual Question Answering (VQA) approaches MCAT explains how histology patches attend to genes, providing an interpretable representation. The method addresses challenges of feature aggregation and data heterogeneity, resulting in superior performance across diverse cancer datasets and is therefore chosen as a baseline.

\paragraph{MM static} was designed to allow for a fair comparison with MM dynamics. To achieve this, both methods share the same structure for the uni-modal encoder. Each encoder learns a latent representation. Before training, weights for each modality can be determined as hyperparameters. These static weights of modalities are multiplied with the latent representation before concatenation.  As in MM dynamics, a classifier is trained on the concatenated input using cross-entropy loss.

\input{Figures/main_image_result_tables}

\subsection{Image MM Dynamics} MM dynamics was originally designed to analyze only tabular genetic data. However, we were able to extend its applicability to image data with some adjustments to two key components of the algorithm. Firstly, we needed to refine how feature informativeness is computed due to the shift in informativeness from individual pixels to groups of neighbouring pixels. It is not meaningful to determine the relevance of individual pixels, as is the case with individual genes. Instead, we grouped neighbouring pixels into patches and determined their significance, which is a more sensible approach. Secondly, we needed to adapt the uni-modal encoder to process image data, which is different from tabular genetic data. The original encoder was not suitable for this purpose.

For the feature informativeness encoder, we chose a compact U-net architecture known for its effectiveness in biomedical segmentation tasks \cite{ronneberger2015unet}. The structure of this model is an encoder-decoder with skip connections between corresponding layers of the encoder and decoder. The encoder uses convolutional and pooling layers to downsample the input image and extract features, while the decoder upsamples the feature maps to produce a segmentation map. The skip connections aid in preserving spatial information and mitigating the vanishing gradient problem during training. 
In our implementation, we used a two-step downsampling process, achieved through convolution and pooling operations. This reduced the input size by a factor of 4 at each step. After each downsampling step, we applied Rectified Linear Unit (ReLU) activation functions to the feature maps.
In contrast, for the decoder, we chose to use a single upsampling convolution operation, omitting the final upsampling convolution and instead using nearest-neighbour interpolation. This method produces a feature informativeness heat-map with patches that are $4\times 4$ pixels in size, ensuring that equal informativeness is given to entire regions rather than individual pixels.
 
For the uni-modal encoder, we use a CNN architecture consisting of two convolutional layers, integrated max-pooling, and two fully connected layers with ReLU activation. This decision was made to create a compact network for proof of concept. However, it is acknowledged that more sophisticated architectures, such as ResNet \cite{he2016deepresnet} or Transformer \cite{vaswani2017attention}, may yield superior outcomes.

%% file: Figures/data_set_figure.tex
\begin{figure}
    \centering
    \begin{subfigure}{0.46\columnwidth} % Specify the width of the first subfigure
        \includegraphics[width=\columnwidth]{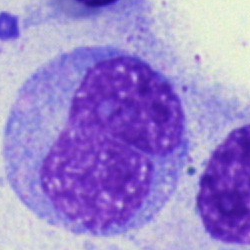} % Change \columnwidth to \linewidth
        \caption{}
    \end{subfigure}
    \hspace{0.02\columnwidth} % Adjust the amount of horizontal space here
    \begin{subfigure}{0.46\columnwidth} % Specify the width of the first subfigure
        \includegraphics[width=\columnwidth]{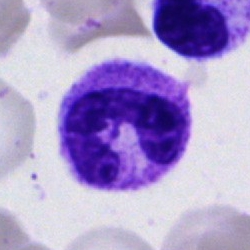} % Change \columnwidth to \linewidth
        \caption{}
    \end{subfigure}\\
    \vspace{0.2cm} % Adjust the amount of horizontal space here
    \begin{subfigure}{0.46\columnwidth} % Specify the width of the first subfigure
        \includegraphics[width=\columnwidth]{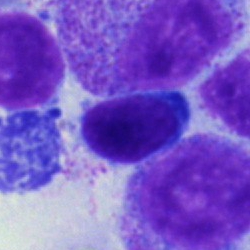} % Change \columnwidth to \linewidth
        \caption{}
    \end{subfigure}
    \hspace{0.02\columnwidth} % Adjust the amount of horizontal space here
    \begin{subfigure}{0.46\columnwidth} % Specify the width of the first subfigure
        \includegraphics[width=\columnwidth]{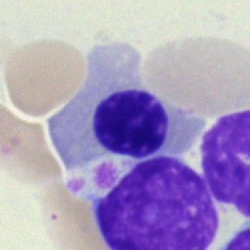} % Change \columnwidth to \linewidth
        \caption{}
    \end{subfigure}
    \caption{Representative examples of the four different cell classes of the data set used in this study \cite{matek2021expert}. (a) displays a monocyte, (b) a neutrophil, (c) a lymphocyte, and (d) an erythroblast.} % Move the \caption command here to caption the entire figure
    \label{fig:dataset_images}
\end{figure}

%% file: Figures/main_results_table_rna_protein.tex
\begin{table*}[ht!]
\centering
\begin{adjustbox}{width=\textwidth}
\begin{tabular}{cccccccc} % Adjust the number of "c" to match the number of columns|
\toprule

Method & Fusion strategy &F1 Score      & F1 macro    & Recall       & Precision     & Accuracy      & Balanced accuracy   \\
\midrule
KNN & early & 94.64 $\pm$ 1.8 & 72.57 $\pm$ 6.7 & 95.36 $\pm$ 1.4 & 95.43 $\pm$ 1.4 & 95.36 $\pm$ 1.4 & 68.06 $\pm$ 6.5 \\
LR & early & \textbf{96.13 $\pm$ 1.1} & \textbf{77.23 $\pm$ 2.4} & \textbf{96.51 $\pm$ 0.9} & \textbf{96.29 $\pm$ 1.1} & \textbf{96.51 $\pm$ 0.9} & \textbf{73.60 $\pm$ 4.1} \\
NN & early & 95.73 $\pm$ 1.3 & 74.58 $\pm$ 2.9 & 96.22 $\pm$ 1.1 & 95.98 $\pm$ 1.3 & 96.22 $\pm$ 1.1 & 71.44 $\pm$ 4.3 \\
RF & early & 94.84 $\pm$ 2.0 & 66.02 $\pm$ 3.2 & 95.89 $\pm$ 1.6 & 94.63 $\pm$ 1.4 & 95.89 $\pm$ 1.6 & 63.72 $\pm$ 3.7 \\
SVM & early & 95.62 $\pm$ 1.9 & 72.42 $\pm$ 5.6 & 96.32 $\pm$ 1.5 & 96.34 $\pm$ 1.5 & \textbf{96.32 $\pm$ 1.5} & 68.79 $\pm$ 5.9 \\
\midrule
HEALNet & intermediate &  95.84 $\pm$ 1.2 & 76.76 $\pm$ 1.6 & 96.22 $\pm$ 1.1 & 95.91 $\pm$ 1.2 & 96.22 $\pm$ 1.1& \textbf{73.58  $\pm$ 0.8}\\
MCAT & late &   95.55  $\pm$ 1.3  &  72.66  $\pm$ 1.6  &  95.99  $\pm$ 1.2  &  95.46  $\pm$  1.2 &  95.99 $\pm$ 1.2  &  70.46 $\pm$ 0.8\\ 
%MM static & late&  94.33  $\pm$ 0.7 &  69.36  $\pm$ 1.2 &  94.71  $\pm$ 0.6 &  94.19  $\pm$0.8  &  94.71  $\pm$ 0.6&  68.99 $\pm$  2.1 \\ 10 1
MM static & late&  94.99  $\pm$ 0.7 & 71.04  $\pm$ 1.2 &  95.53  $\pm$ 0.6 &  94.89  $\pm$ 0.8  &  95.52  $\pm$ 0.6&  69.12 $\pm$  2.1 \\ % 3 1
\midrule
MM dynamics  & late & 94.32 $\pm$ 1.7& 66.67 $\pm$ 2.7 & 95.13 $\pm$ 1.5& 93.90 $\pm$ 1.7& 95.13 $\pm$ 1.5& 64.53 $\pm$ 2.6\\
%MM dynamics 2000 epochs & &95.33 $\pm$ 1.71 & 71.21 $\pm$ 1.37 & 96.07 $\pm$ 1.51 & 95.68 $\pm$ 1.72 & 96.07 $\pm$ 1.51 & 67.61 $\pm$ 2.26 \\ 
\bottomrule
\end{tabular}
\end{adjustbox}
\caption{Comparison of state-of-the-art methods using the modalities RNA and protein. 
The best results per column are in bold. The analysis reveals that basic machine learning classifiers exhibit the highest performance, particularly logistic regression (LR). Among deep learning models, only HEALNet achieves partially comparable results.}
\label{tab:main_results_rna_protein}
\end{table*} 
\begin{figure*}
    \centering
    \includegraphics[width=\textwidth]{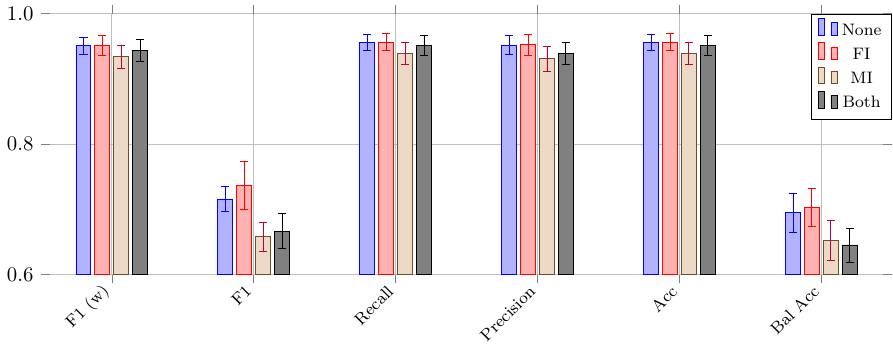}
    \caption{Overview of ablation studies that examine the impact of feature informativeness (FI) and modality informativeness (MI) components in the MM dynamics approach. FI refers to the inclusion of feature informativeness, MI refers to the inclusion of modality informativeness, and 'Both' indicates the incorporation of both FI and MI components. 'None' denotes scenarios where neither FI nor MI components are utilised. The figure displays the results across various evaluation metrics. All results were generated using MM dynamics trained on RNA and protein with latent representation dimensions of 250 and 35, respectively.}
    \label{fig:informativnessablation}
\end{figure*}

%%%%%%%%%%%

%% file: Figures/main_image_result_tables.tex
\begin{table*}[htbp]
\centering
\begin{adjustbox}{width=\textwidth}
\begin{tabular}{cccccccc} % Adjust the number of "c" to match the number of columns|
\toprule

Method  & Modalities&F1 Score      & F1 macro    & Recall       & Precision     & Accuracy      & Balanced accuracy   \\
\midrule
CNN &  I & 87.00  $\pm$ 1.7 &  63.00  $\pm$ 2.1 &  85.20  $\pm$ 1.4 &  89.75  $\pm$ 1.5 &  85.20  $\pm$ 1.5&  73.03 $\pm$ 1.7 \\

\midrule

%image (None: MI) &  84.15  $\pm$ 0.2 &  58.91  $\pm$ 2.9&  81.34  $\pm$ 0.5 &  88.48  $\pm$ 1.1&  81.34  $\pm$ 0.5 &  70.67 $\pm$ 2.1\\

%tree (None: MI) &  95.29  $\pm$ 1.7 &  73.81  $\pm$ 4.3 &  95.83  $\pm$ 1.5 &  95.44  $\pm$ 1.5  &  95.83  $\pm$ 1.5 &  70.19  $\pm$ 4.2  

 MM dynamics& R + I &   95.54  $\pm$ 1.3  &  74.35  $\pm$  3.4 &  96.12  $\pm$  1.2 &  95.82  $\pm$  1.4 &  96.12  $\pm$  1.2 &  70.07  $\pm$  2.5 \\ 

 MM dynamics & P + I &83.55 $\pm$ 0.8 & 60.36 $\pm$ 5.7 & 81.14 $\pm$ 2.2 & 87.48 $\pm$ 1.4 & 81.14 $\pm$ 2.2& 71.67 $\pm$ 2.7\\

MM static &P + R + I &  \textbf{96.74  $\pm$ 0.7 }& \textbf{ 81.01  $\pm$ 2.3} &  \textbf{97.03  $\pm$ 0.6}&  \textbf{96.84  $\pm$ 0.6} &  \textbf{97.03  $\pm$ 0.6} &  \textbf{76.97  $\pm$ 0.4} \\  
\midrule
 MM dynamics  &P + R + I &  95.98 $\pm$ 1.7 &   78.64 $\pm$ 2.1 &   96.50 $\pm$ 1.4  &   96.41 $\pm$ 1.4  &   96.50 $\pm$ 1.7 &   73.50 $\pm$  1.7  \\

\bottomrule
\end{tabular}
\end{adjustbox}
\caption{Overview of the results obtained using new Image MM dynamics extension. For comparison, MM static was also calculated as a baseline on all three modalities, along with ablations using Image MM dynamics only on image and another modality, as well as solely image for comparison.  The results show that while Image MM dynamics cannot match MM static and the latter remains superior, there is a significant improvement compared to using only two modalities. However, the use of three modalities resulted in a significant improvement compared to using only two, indicating the effectiveness of integrating images into the original MM dynamics algorithm. The abbreviations used are R for RNA, P for protein, and I for images with latent representation dimensions of 250, 35 and 500, respectively.}
\label{tab:image_results}
\end{table*} 
%%%%%%%
\begin{table*}[ht!]
\centering
\begin{adjustbox}{width=\textwidth}
\begin{tabular}{cccccccc} % Adjust the number of "c" to match the number of columns|
\toprule

Method  & Modalities&F1 Score      & F1 macro    & Recall       & Precision     & Accuracy      & Balanced accuracy   \\
\midrule

%image (None: MI) &  84.15  $\pm$ 0.2 &  58.91  $\pm$ 2.9&  81.34  $\pm$ 0.5 &  88.48  $\pm$ 1.1&  81.34  $\pm$ 0.5 &  70.67 $\pm$ 2.1\\

%tree (None: MI) &  95.29  $\pm$ 1.7 &  73.81  $\pm$ 4.3 &  95.83  $\pm$ 1.5 &  95.44  $\pm$ 1.5  &  95.83  $\pm$ 1.5 &  70.19  $\pm$ 4.2  \\
MM dynamics& P + R + I* & 92.97 $\pm$ 3.8& 64.69 $\pm$ 4.8& 94.19 $\pm$ 3.1& 93.78 $\pm$ 2.3& 94.19 $\pm$ 3.1& 61.58 $\pm$ 4.8\\
MM dynamics & P + R & 94.32 $\pm$ 1.7& 66.67 $\pm$ 2.7 & 95.13 $\pm$ 1.5& 93.90 $\pm$ 1.7& 95.13 $\pm$ 1.5& 64.53 $\pm$ 2.6\\

MM static & P + R + I* &  93.82 $\pm$ 2.9 & 70.65 $\pm$ 4.9 & 94.43 $\pm$2.6 & 94.17 $\pm$ 2.5 & 94.43 $\pm$ 2.6& 66.54 $\pm$ 4.9\\ 

\bottomrule
\end{tabular}
\end{adjustbox}
\caption{Overview of the results obtained using new image MM dynamics approaches when masking the image modality. The images were masked only during testing by substituting them with gray images of uniform intensity 0.5. Additionally, MM static was included as a baseline for comparison. The results indicate that MM dynamics performs noticeably worse than the static late fusion baseline or dynamic late fusion trained solely on the available modalities. The abbreviations used are R for RNA, P for protein, and I* for masked images with latent representation dimensions of 250, 35 and 500, respectively.}
\label{tab:image_results_mask}
\end{table*} 
%%%%%

%% file: sections/04_Results.tex
\section{Results} 
\label{section:results}

The following section highlights the results of quantitative and qualitative analyses of the established approach in Subsection \ref{subsec:pureoldmmdynamics}. We then explore the implications and findings of the novel extension of Image MM dynamics, providing a comprehensive account of its performance in Subsection \ref{subsec:fancynewimagedynamcs}. Further details on results not presented in this section can be found in Appendix \ref{sec:appendix}.

Table \ref{tab:main_results_rna_protein} summarises the performance metrics obtained from experiments on tabular genetic data using the modalities RNA and protein.  Our experiements show that MM dynamics is less effective than traditional classifiers and lags behind intermediate and late fusion state-of-the-art methods. LR is the most effective classifier, demonstrating superior performance in mitigating challenges associated with large class imbalances, as evidenced by good performance on metrics such as F1 macro and balanced accuracy. Of all the deep neural network approaches, only HEALNet achieves results comparable to those of LR. However, MM dynamics face challenges in generalising to smaller classes and notably underperform on balanced accuracy and F1 macro. Contrary to expectations, the results do not confirm MM dynamics' supposed reduction of irrelevant information through dynamic fusion.

\begin{figure}
    \centering
    \includegraphics[width=\columnwidth]{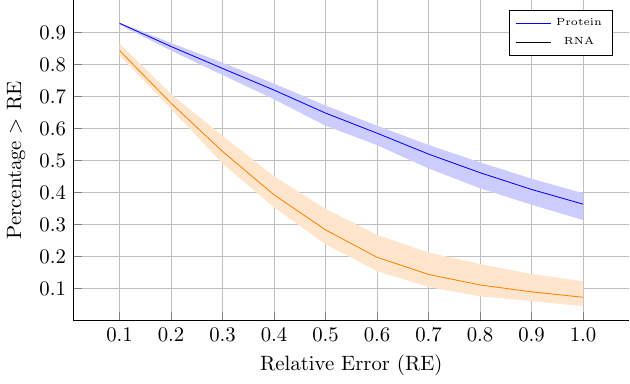}
    \caption{Overview of classification confidence. The chart shows the relative error of $TCP$. The x-axis represents the threshold of the relative error. The y-axis shows the percentage of instances with an RE greater than $x$: $x\cdot {TCP} + {TCP} \leq \widehat{T C P}$. $TCP$ was extracted from the best-performing MM dynamics network with latent dimensions of 35 for proteins and 250 for RNA. }
    \label{fig:tcp relative}
\end{figure}

\subsection{Analysis of MM Dynamics Components} 
\label{subsec:pureoldmmdynamics}
The following analysis will investigate the feature and modality informativeness components of the MM dynamics algorithm due to the lack of performance reproducibility.

\paragraph{Impact of Feature and Modality Informativeness}
The study further examined the impact of the feature informativeness (FI) and modality informativeness (MI) components in the MM dynamics approach. Figure \ref{fig:informativnessablation} compares the FI and MI components, showing that the inclusion of the feature informativeness component does not result in a noticeable improvement, except for the balanced metrics across all classes (balanced accuracy, F1 macro), where an improvement can be seen. It can be concluded that FI is useful in addressing class imbalances. In contrast, when MI is used, a significant decline is observed, particularly in the balanced metrics. This indicates that MI does not effectively reduce noise but rather introduces noise, leading to a bias towards predicting overrepresented classes. Combining both components produces slightly better results than MI alone but is still inferior to the scenario with no informativeness components. Therefore, it can be inferred that only FI is relevant and effective in improving model performance, and MI is unsuitable.

%\widehat{T C P}
%{T C P}
\paragraph{Classification Confidence Analysis}
\begin{figure}
    \centering
    \begin{subfigure}{0.45\textwidth} % Specify the width of the second subfigure
        \includegraphics[width=\textwidth]{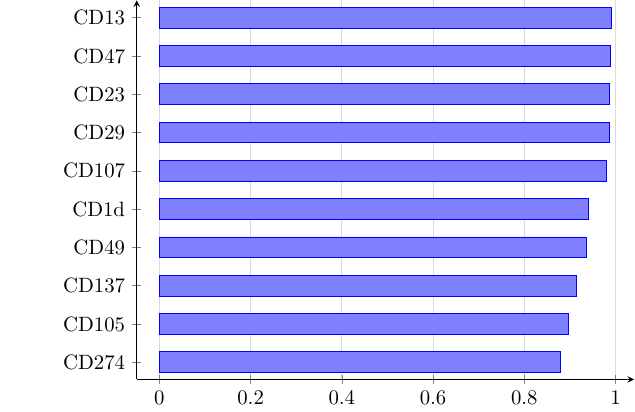} % Change \columnwidth to \linewidth
        \caption{Protein}
    \end{subfigure}
    \hfill % Add horizontal space between subfigures
    \begin{subfigure}{0.45\textwidth} % Specify the width of the first subfigure
        \includegraphics[width=\textwidth]{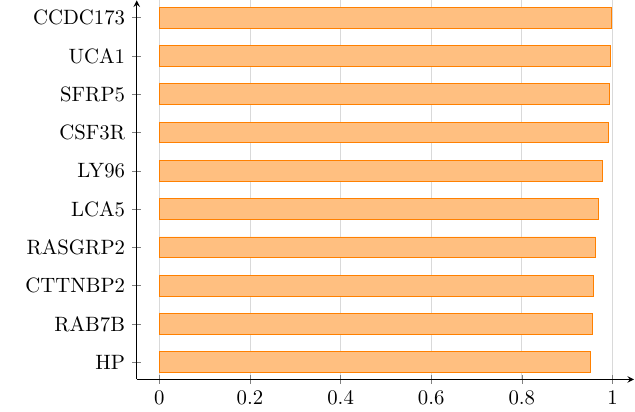} % Change \columnwidth to \linewidth
        \caption{RNA}
    \end{subfigure}

    \caption{The top 10 informative biomarkers are displayed from the modalities protein (b) and RNA (b) identified by MM dynamics feature informativeness encoder.} % Move the \caption command here to caption the entire figure
    \label{fig:mostimportenfeatures}
\end{figure}

Two potential causes were considered when investigating the reasons for the ineffectiveness of the modality informativeness (MI) component. Firstly, there may be no relationship between the classification confidence and modality informativeness, or, secondly, such a relationship cannot be accurately predicted. 
Therefore, an analysis of the prediction of the classification confidence termed true class probability $({T C P})$, was conducted. The mean absolute error between the true $TCP$ and estimated $\widehat{T C P}$ for protein was $0.36 \pm 0.37$, and for RNA, it was $0.35 \pm 0.33$, with a maximum error of around $3.23$ and $3.53$, respectively.
The  ${T C P}$ ranges between $[0,1]$, where a $\mu({T C P})$ close to 1 is indicative of a reliable and secure uni-modal classifier. However, the $\mu({T C P})$ for protein was  $0.447$, and for RNA, it was $0.471$. This suggests that the confidence loss, as defined in Equation \ref{eq:confidenceloss}, not only adjusts the estimated $\widehat{T C P}$ but also leads to the model making less confident decisions, resulting in a decrease in the actual $TCP$ values. Therefore, relying solely on the mean absolute error may not accurately assess model performance.

We evaluated the errors relative to the true $TCP$ to gain further insights. Figure \ref{fig:tcp relative} provides an overview, with the x-axis representing the relative error: $x\cdot {TCP} + {TCP} \leq \widehat{T C P}$. The y-axis denotes the percentage of instances with a relative error (RE) greater than $x$.
It is evident that the RE frequently exceeds the threshold of $0.1$. For the modality protein, over 90\% of instances in the test dataset exhibit an RE higher than $0.1$, while for RNA, this figure is around 85\%. Importantly, approximately 50\% of protein instances have an RE of approximately $0.7$, while for RNA, it is around $0.3$. Around 35\% instances of the modality protein show an estimated $TCP$ that deviates by 100\% from the actual $TCP$, indicating a failure in predicting $\widehat{T C P}$ and explaining the lack of success in the modality informativeness component. It is still an open research question whether there is a relationship between classification confidence and modality informativeness.

\begin{figure}
    \centering
    \begin{subfigure}{0.48\textwidth} % Specify the width of the first subfigure
        \includegraphics[width=\textwidth]{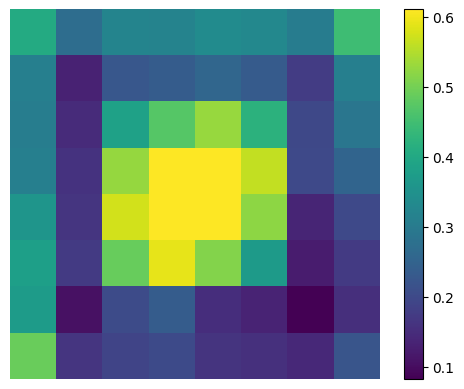} % Change \columnwidth to \linewidth
    \end{subfigure}
    \caption{Heat-map displaying the mean feature informativeness across all test data images. The results show that the central regions of the images, where the cell is usually located, are considered the most relevant.} % Move the \caption command here to caption the entire figure
    \label{fig:imagemmdynamics}
\end{figure}

\paragraph{Missing Modalities} Furthermore, we examined whether modality informativeness alone can effectively suppress missing modalities that have been replaced by noise, thereby preserving classification performance. 
Table \ref{tab:image_results_mask} presents the results of using only protein and RNA modalities during training versus using all three modalities for training, with gray images substituted by uniform intensity of 0.5 during testing. 
Performance deteriorates when the missing modality is masked during testing, particularly in balanced accuracy, which decreases by 3 percentage points. It is noteworthy that MM static outperforms MM dynamics even when images are masked, indicating that MM dynamics is unsuitable for applications with regular occurrences of missing modalities.

%B Zellen: CD19, CD20, CD21 und CD40.
% monozyteb; CD14, CD11b, CD68, EMR1, Lysozym M oder MAC-1/MAC-3.
% Erythroblast: GPA, RhAG und CD47. 

\paragraph{Biomarker identification} 
\begin{figure}
    \centering
    \begin{subfigure}{0.2\textwidth} % Specify the width of the first subfigure
        \includegraphics[width=\textwidth]{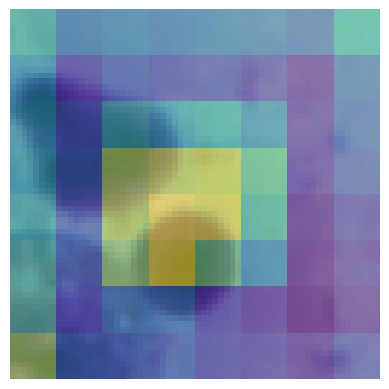} % Change \columnwidth to \linewidth
       
    \end{subfigure}
    \begin{subfigure}{0.2\textwidth} % Specify the width of the first subfigure
        \includegraphics[width=\textwidth]{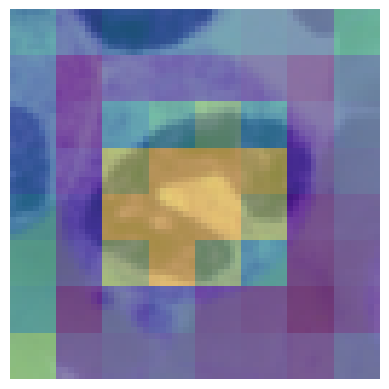} % Change \columnwidth to \linewidth
      
    \end{subfigure}
    \hfill % Add horizontal space between subfigures
    \begin{subfigure}{0.2\textwidth} % Specify the width of the first subfigure
        \includegraphics[width=\textwidth]{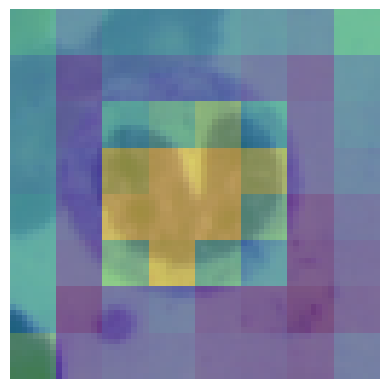} % Change \columnwidth to \linewidth
       
    \end{subfigure}
    \begin{subfigure}{0.2\textwidth} % Specify the width of the first subfigure
        \includegraphics[width=\textwidth]{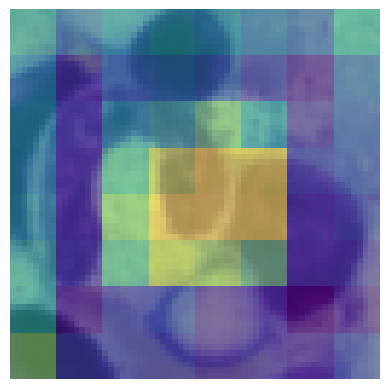} % Change \columnwidth to \linewidth
        
    \end{subfigure}
    \caption{Example images from the test dataset overlaid with the feature informativeness heat-maps generated per image reveal a concentration of relevance towards the centre of the image, even when multiple cells are visible.} % Move the \caption command here to caption the entire figure
    \label{fig:heat_maps_feature_inforativenes}
\end{figure}

In our case, we have classified cell types rather than cancer subtypes. It is, therefore, reasonable to assume that relevant surface proteins and RNA would be identified for each cell type.
Figure \ref{fig:mostimportenfeatures} illustrates the top 10 genetic features per modality. The most important protein is CD13 which is recognised as a moonlighting enzyme owing to its involvement in multiple functions \cite{mina2008moonlightingcd13}. CD13 is expressed as a surface protein across all four cell classes examined in this investigation \cite{young1999humanc13}. Furthermore, a growing interest has been in utilising CD13 as a therapeutic target \cite{wickstrom2011aminopeptidasecd13}. Similarly, CD47 is a target for cancer immunotherapy \cite{jiang2021targetingcd47} and is also observed as a surface protein on erythroblasts. For instance, CD23 is a marker for B-lymphocytes \cite{delespesse1991expressioncd23}. These examples show that surface proteins crucial for classification were learned and surface proteins targeted by cancer therapies were identified.

%\paragraph{Density of feature informativeness}
%\paragraph{Density of modality informativeness}

\subsection{Analysis of Image MM Dynamics} 
\label{subsec:fancynewimagedynamcs}
The evaluation of the novel Image MM dynamics extension focuses primarily on the feature informativeness encoder. First, qualitative results are discussed, followed by the presentation of feature informativeness heat maps. 
Table \ref{tab:image_results} shows the results of Image MM dynamics. A significant improvement can be seen, particularly in F1 macro and balanced accuracy, when training with all three modalities compared to only protein and RNA. In addition, Table \ref{tab:image_results} shows results when using RNA and images only, or protein and images, as well as images only. A clear improvement can be seen in all configurations, particularly when compared to the image-only baseline, which showed an increase from 63\% $\pm$ 2.1 F1 macros to 78.6\% $\pm$ 2.1 by incorporating the modalities protein and RNA in addition to images. Unfortunately, there is still no improvement over the baseline approach MM static. This still underlines the success of integrating novel feature informativeness encoders as the lower performance can be attributed to the MI components from MM dynamics, as previously discussed.
The heat-maps in Figure  \ref{fig:imagemmdynamics} and Figure \ref{fig:heat_maps_feature_inforativenes} provide insight into the functionality of the feature informativeness encoder. Figure \ref{fig:imagemmdynamics} shows the mean feature informativeness, which highlights a clear focus on the center of the image. This section corresponds to the area containing the important single cell, indicating successful learning of relevant features, as confirmed by Figure \ref{fig:heat_maps_feature_inforativenes}. However, additional weight has been assigned to the edges and corners, which appears unwarranted and lacks explanation.
Overall, our proposed image encoder structure has qualitatively and quantitatively demonstrated functionality, validating its effectiveness.

%% file: sections/05_Discussion.tex
\section{Conclusion}
\label{section:discussion}
This paper introduces a new extension of MM dynamics proposed by \citet{Han_Yang_Huang_Zhang_Yao_2022} termed Image MM dynamics, which incorporates explainable image classification. The initial aim was to replicate the results of MM dynamics on a new dataset. However, despite the promising results reported in the original paper, the results could not be reproduced on the new data set.

Modality informativeness could not be confirmed as a viable fusion strategy, as it could not be predicted sufficiently well, resulting in consistent underperformance compared to the baseline MM static.
However, the inclusion of feature informativeness improved performance and enhanced explainability. As a result, we expanded this component of MM dynamics to images, resulting in the development of Image MM dynamics. The qualitative results demonstrate the success of this extension. Furthermore, the integrated explainability demonstrates that the relevant area for classification has been learned.
However, further testing is required on images where the target varies across samples throughout the image rather than being solely located in the centre. 
Moreover, further exploration can involve testing more precise model architectures, such as deeper U-Nets, to enhance feature informativeness segmentation.

In conclusion, the feature informativeness in MM dynamics presents a promising opportunity to improve explainability and potentially enhance performance. This has been demonstrated in different domains, including image and tabular data.

%'Overfitting to the training data, and therefore poor generalizability, is a major challenge for multimodal models' \cite{Stahlschmidt_Ulfenborg_Synnergren_2022}

%-  mit dem most importand word und der auswertung: keine globale auswertung mehr möglich bei bildern und text, da jede zelle für etwas anderses steht per sample

%% file: sections/09_Appendix.tex
The appendix includes important ablation studies that were necessary to reproduce the results. The original paper did not publish hyperparameters, so we tested the optimal hyperparameters for the dimensions of the latent representation. Table \ref{tab:hidden dim} shows that there is no significant difference between the dimensions ranging from 35 to 250 compared to 140 and 1000. Therefore, we decided to use the lower dimensionality.

\begin{table*}[ht!]
\centering
\begin{adjustbox}{width=\textwidth}
\begin{tabular}{cccccccc} % Adjust the number of "c" to match the number of columns|
\toprule

Hidden Dimensions & F1 Score & F1 Macro & Recall & Precision & Accuracy & Balanced Accuracy \\
\midrule
 P: 35, R: 250  &  \textbf{94.32 $\pm$ 1.74}& \textbf{66.67 $\pm$ 2.75} & 95.13 $\pm$ 1.58& 93.90 $\pm$ 1.73& \textbf{95.13 $\pm$ 1.57}& \textbf{64.53 $\pm$ 2.65} \\
P: 70, R: 500 &  93.92 $\pm$ 1.90  &  65.56 $\pm$ 2.71  &  94.97 $\pm$ 1.66  &  93.91 $\pm$ 1.63  &  94.97 $\pm$ 1.66  &  63.75 $\pm$ 3.25    \\
P: 140, R: 1000 &  94.28 $\pm$ 1.89  &  65.61 $\pm$ 2.29  &  \textbf{95.18 $\pm$ 1.67 } &  \textbf{94.17 $\pm$ 2.49}  &  95.18 $\pm$ 1.67  &  63.46 $\pm$ 1.96   \\

\bottomrule

\end{tabular}
\end{adjustbox}
\caption{Overview of results using different latent representation dimensions. Each model was trained for 250 epochs under identical conditions. The abbreviations used are R for RNA, P for protein.}
\label{tab:hidden dim}
\end{table*} 

We trained the model using varying lambda parameters for the different loss functions. The results are presented in Table \ref{tab:loss importance }. The performance degrades when the modality informativeness loss or feature informativeness loss is upregulated, but there is no significant change when the classification loss is increased. Therefore, we decided to maintain equal weighting for all losses, consistent with the original implementation.

% \lambda_1 \mathcal{L}^{\ell_1}+\lambda_2 \mathcal{L}^{\text {conf}}+\lambda_3 \mathcal{L}^f\right)
\begin{table*}[ht!]
\centering
\begin{adjustbox}{width=\textwidth}
\begin{tabular}{cccccccccc} % Adjust the number of "c" to match the number of columns|
\toprule

$ \lambda_1 \mathcal{L}^{\ell_1} $& $\lambda_2\mathcal{L}^{\text {conf}}$ & $\lambda_3 \mathcal{L}^f$ & F1 Score & F1 Macro & Recall & Precision & Accuracy & Balanced Accuracy \\
\midrule
  1  & 1& 1& 94.23  $\pm$ 1.91  &  68.53  $\pm$ 4.31  &  94.92  $\pm$ 1.63  &  94.08  $\pm$ 2.16  &  94.92  $\pm$ 1.63  &  66.57  $\pm$ 4.44   \\

 10  & 1& 1  &   94.34  $\pm$ 1.72  &  67.02  $\pm$ 1.66  &  95.15  $\pm$ 1.59  &  94.11  $\pm$ 1.46  &  95.15  $\pm$ 1.59  &  64.65  $\pm$ 2.24   \\
 1  & 10& 1& 94.32   $\pm$  1.74& 66.67   $\pm$ 2.75 & 95.13   $\pm$  1.58& 93.90   $\pm$  1.73& 95.13   $\pm$  1.57& 64.53   $\pm$  2.65\\
  1  & 1& 10&  94.25  $\pm$ 1.70  &  68.58  $\pm$ 4.07  &  94.91  $\pm$ 1.40  &  94.16  $\pm$ 1.99  &  94.91  $\pm$ 1.40  &  66.80  $\pm$ 4.59   \\
   
\bottomrule

\end{tabular}
\end{adjustbox}
\caption{Overview of the results obtained using different weightings for the three components of the loss function. Specifically, the latent representation dimensions were set to 35 and 250 for the protein and RNA modalities, respectively. The results demonstrate variations in performance metrics based on the weighting assigned to each part of the loss function. }
\label{tab:loss importance }
\end{table*}

\begin{figure*} [!htbp]
    \centering
    \includegraphics[width=\textwidth]{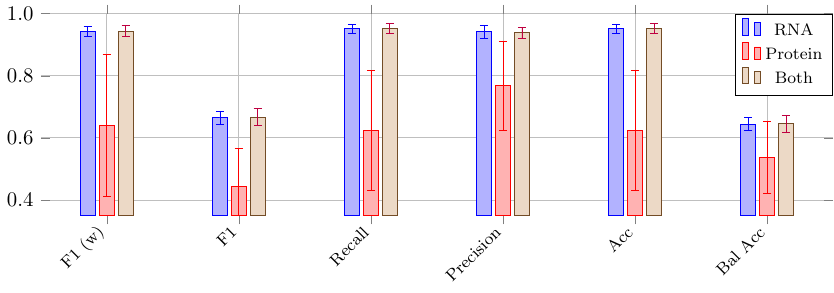}
    \caption{Result of the ablation of the modalities including both genetic table data using MM dynamic as a model.}
    \label{fig:ablation_protein_rna}
\end{figure*}

In addition, ablation studies were conducted for the two genetic tabular modalities. The results for MM dynamics are shown in Figure 
\ref{fig:ablation_protein_rna}. It is clear that using only RNA as a modality is more effective than protein. The integration of both does not lead to an improvement. Tables \ref{tab:results_rna} and \ref{tab:results_protein} present the results of RNA and protein, respectively, compared to baseline ML classifiers. LR and NN outperform MM dynamics in terms of performance. This is particularly evident in protein classification, where LR achieves an F1 macro score of 76.5\%, while MM dynamics only achieve a score of 44.5\%.

\begin{table*}[!htbp]
\centering
\begin{adjustbox}{width=\textwidth}
\begin{tabular}{cccccccc} % Adjust the number of "c" to match the number of columns|
\toprule

Method & Fusion strategy &F1 Score      & F1 macro    & Recall       & Precision     & Accuracy      & Balanced accuracy   \\
\midrule
KNN & Early & 89.45 $\pm$ 2.1 & 60.20 $\pm$ 3.1 & 90.76 $\pm$ 1.7 & 90.27 $\pm$ 1.5 & 90.76 $\pm$ 1.7 & 56.77 $\pm$ 1.9 \\
LR & Early & \textbf{94.39 $\pm$ 1.6 }& \textbf{70.76 $\pm$ 2.0} & 94.86 $\pm$ 1.4 & \textbf{94.48 $\pm$ 1.7} & 94.86 $\pm$ 1.4 &\textbf{ 68.65 $\pm$ 2.7} \\
NN & Early &\textbf{ 94.38 $\pm$ 1.7} & \textbf{70.61 $\pm$ 2.6} & \textbf{95.02 $\pm$ 1.4} & \textbf{94.50 $\pm$ 1.4} &\textbf{ 95.02 $\pm$ 1.4} & 67.62 $\pm$ 3.5 \\
RF & Early & 93.34 $\pm$ 2.5 & 61.96 $\pm$ 3.3 & 94.75 $\pm$ 2.0 & 93.60 $\pm$ 1.7 & 94.75 $\pm$ 2.0 & 59.11 $\pm$ 2.6 \\
SVM & Early & 94.48 $\pm$ 1.8 & 65.91 $\pm$ 2.6 & 95.48 $\pm$ 1.5 & 94.18 $\pm$ 1.4 & 95.48 $\pm$ 1.5 & 63.46 $\pm$ 2.7 \\
\midrule
MM dynamics  & late &    94.29 $\pm$ 1.6 &  66.42 $\pm$ 2.2 &  95.15 $\pm$ 1.4 &  94.2 $\pm$ 2.15 &  95.15 $\pm$ 1.4 &  64.42 $\pm$ 2.1 \\
\bottomrule
\end{tabular}
\end{adjustbox}
\caption{ Comparison of state-of-the-art methods using the modality RNA. }
\label{tab:results_rna}
\end{table*} 

\begin{table*}[!htbp]
\centering
\begin{adjustbox}{width=\textwidth}
\begin{tabular}{cccccccc} % Adjust the number of "c" to match the number of columns|
\toprule

Method & Fusion strategy &F1 Score      & F1 macro    & Recall       & Precision     & Accuracy      & Balanced accuracy   \\
\midrule
KNN & Early & 94.56 $\pm$ 1.2 & 73.27 $\pm$ 2.8 & 95.10 $\pm$ 1.1 & 94.74 $\pm$ 1.0 & 95.10 $\pm$ 1.1 & 69.67 $\pm$ 3.4 \\
LR & Early &\textbf{94.93 $\pm$ 0.9} &\textbf{76.45 $\pm$ 1.8} & 95.21 $\pm$ 0.9 & 94.90 $\pm$ 0.8 & 95.21 $\pm$ 0.9 & \textbf{73.69 $\pm$ 3.6} \\
NN & Early & \textbf{95.02 $\pm$ 0.9} & 75.32 $\pm$ 2.8 & 95.29 $\pm$ 1.1 & 95.04 $\pm$ 0.9 & 95.29 $\pm$ 1.1 & 72.96 $\pm$ 2.7 \\
RF & Early & 94.31 $\pm$ 1.6 & 66.41 $\pm$ 2.8 & 95.25 $\pm$ 1.3 & 94.04 $\pm$ 1.1 & 95.25 $\pm$ 1.3 & 64.53 $\pm$ 3.7 \\
SVM & Early & 95.05 $\pm$ 1.2 & 70.42 $\pm$ 2.2 & \textbf{95.77 $\pm$ 1.0} & \textbf{95.46 $\pm$ 1.2} & \textbf{95.77 $\pm$ 1.0} & 67.62 $\pm$ 3.7 \\
\midrule
MM dynamics  & late &  64.07 $\pm$  22.9 & 44.45 $\pm$ 12.0 & 62.41 $\pm$  19.2 & 76.78 $\pm$ 14.3 & 62.41 $\pm$ 19.2 & 53.58 $\pm$  11.6 \\
\bottomrule
\end{tabular}
\end{adjustbox}
\caption{Comparison of state-of-the-art methods using the modality protein.}
\label{tab:results_protein}
\end{table*}

Furthermore, we conducted ablation studies on the informativeness components, as already illustrated in Figure \ref{fig:informativnessablation}. Detailed numerical results, including the mean and standard deviation, are provided in Table \ref{tab: importance of informativness}.
\begin{table*}[!htbp]
\centering
\begin{adjustbox}{width=\textwidth}
\begin{tabular}{cccccccc} % Adjust the number of "c" to match the number of columns|
\toprule

Loss & F1 Score & F1 Macro & Recall & Precision & Accuracy & Balanced Accuracy \\
\midrule
None  &   95.09 $\pm$ 1.33&  71.59 $\pm$ 1.99 &  95.59 $\pm$ 1.23 &  95.14 $\pm$ 1.46 
 &  95.59 $\pm$ 1.23 &  69.46 $\pm$ 3.01\\
FI   &  \textbf{95.13 $\pm$ 1.50} &  \textbf{73.66 $\pm$ 3.69} &  \textbf{95.64 $\pm$ 1.35}&  \textbf{95.23 $\pm$ 1.61}&  \textbf{95.64  $\pm$ 1.35}&  \textbf{70.25 $\pm$ 2.89}\\
MI& 93.40 $\pm$ 1.78 & 65.77 $\pm$ 2.24 & 93.86 $\pm$1.67  & 93.09 $\pm$  1.88& 93.86 $\pm$ 1.67 & 65.17 $\pm$  3.06 \\

\bottomrule

\end{tabular}
\end{adjustbox}
\caption{Overview of ablation studies that examine the impact of feature informativeness (FI) and modality informativeness (MI) components in the MM dynamics approach. }
\label{tab: importance of informativness}
\end{table*}

%% file: main.bbl
\begin{thebibliography}{21}
\expandafter\ifx\csname natexlab\endcsname\relax\def\natexlab#1{#1}\fi

\bibitem[{Acosta et~al.(2022)Acosta, Falcone, Rajpurkar, and Topol}]{Acosta_Falcone_Rajpurkar_Topol_2022}
Julián~N. Acosta, Guido~J. Falcone, Pranav Rajpurkar, and Eric~J. Topol. 2022.
\newblock \href {https://doi.org/10.1038/s41591-022-01981-2} {Multimodal biomedical ai}.
\newblock \emph{Nature Medicine}, 28(9):1773–1784.

\bibitem[{Azam et~al.(2022)Azam, Ryabchykov, and Bocklitz}]{Azam_Ryabchykov_Bocklitz_2022}
Kazi Sultana~Farhana Azam, Oleg Ryabchykov, and Thomas Bocklitz. 2022.
\newblock \href {https://doi.org/10.3390/molecules27217448} {A review on data fusion of multidimensional medical and biomedical data}.
\newblock \emph{Molecules}, 27(21):7448.

\bibitem[{Burkhardt et~al.(2022)Burkhardt, Luecken, Benz, Holderrieth, Bloom, Lance, Chow, and Holbrook}]{open-problems-multimodal}
Daniel Burkhardt, Malte Luecken, Andrew Benz, Peter Holderrieth, Jonathan Bloom, Christopher Lance, Ashley Chow, and Ryan Holbrook. 2022.
\newblock \href {https://kaggle.com/competitions/open-problems-multimodal} {Open problems - multimodal single-cell integration}.

\bibitem[{Chen et~al.(2021)Chen, Lu, Weng, Chen, Williamson, Manz, Shady, and Mahmood}]{Chen_Lu_Weng_Chen_Williamson_Manz_Shady_Mahmood_2021mcat}
Richard~J. Chen, Ming~Y. Lu, Wei-Hung Weng, Tiffany~Y. Chen, Drew~Fk. Williamson, Trevor Manz, Maha Shady, and Faisal Mahmood. 2021.
\newblock \href {https://doi.org/10.1109/ICCV48922.2021.00398} {Multimodal co-attention transformer for survival prediction in gigapixel whole slide images}.
\newblock In \emph{2021 IEEE/CVF International Conference on Computer Vision (ICCV)}, page 3995–4005, Montreal, QC, Canada. IEEE.

\bibitem[{Delespesse et~al.(1991)Delespesse, Suter, Mossalayi, Bettler, Sarfati, Hofstetter, Kilcherr, Debre, and Dalloul}]{delespesse1991expressioncd23}
G~Delespesse, U~Suter, D~Mossalayi, B~Bettler, M~Sarfati, H~Hofstetter, E~Kilcherr, P~Debre, and A~Dalloul. 1991.
\newblock Expression, structure, and function of the cd23 antigen.
\newblock \emph{Advances in immunology}, 49:149--191.

\bibitem[{Dimitrovski et~al.(2015)Dimitrovski, Kocev, Kitanovski, Loskovska, and Džeroski}]{Dimitrovski_Kocev_Kitanovski_Loskovska_Džeroski_2015}
Ivica Dimitrovski, Dragi Kocev, Ivan Kitanovski, Suzana Loskovska, and Sašo Džeroski. 2015.
\newblock \href {https://doi.org/10.1016/j.compmedimag.2014.06.005} {Improved medical image modality classification using a combination of visual and textual features}.
\newblock \emph{Computerized Medical Imaging and Graphics}, 39:14–26.

\bibitem[{Han et~al.(2022)Han, Yang, Huang, Zhang, and Yao}]{Han_Yang_Huang_Zhang_Yao_2022}
Zongbo Han, Fan Yang, Junzhou Huang, Changqing Zhang, and Jianhua Yao. 2022.
\newblock \href {https://doi.org/10.1109/CVPR52688.2022.02005} {Multimodal dynamics: Dynamical fusion for trustworthy multimodal classification}.
\newblock In \emph{2022 IEEE/CVF Conference on Computer Vision and Pattern Recognition (CVPR)}, page 20675–20685, New Orleans, LA, USA. IEEE.

\bibitem[{He et~al.(2016)He, Zhang, Ren, and Sun}]{he2016deepresnet}
Kaiming He, Xiangyu Zhang, Shaoqing Ren, and Jian Sun. 2016.
\newblock Deep residual learning for image recognition.
\newblock In \emph{Proceedings of the IEEE conference on computer vision and pattern recognition}, pages 770--778.

\bibitem[{Hemker et~al.(2023)Hemker, Simidjievski, and Jamnik}]{Hemker_Simidjievski_Jamnik_2023HEALNet}
Konstantin Hemker, Nikola Simidjievski, and Mateja Jamnik. 2023.
\newblock \href {http://arxiv.org/abs/2311.09115} {Healnet -- hybrid multi-modal fusion for heterogeneous biomedical data}.
\newblock (arXiv:2311.09115).

\bibitem[{Jiang et~al.(2021)Jiang, Sun, Yu, Tian, and Song}]{jiang2021targetingcd47}
Zhongxing Jiang, Hao Sun, Jifeng Yu, Wenzhi Tian, and Yongping Song. 2021.
\newblock Targeting cd47 for cancer immunotherapy.
\newblock \emph{Journal of hematology \& oncology}, 14(1):180.

\bibitem[{Matek et~al.(2021)Matek, Krappe, Münzenmayer, Haferlach, and Marr}]{matek2021expert}
Christian Matek, Sebastian Krappe, Christian Münzenmayer, Torsten Haferlach, and Carsten Marr. 2021.
\newblock \href {https://doi.org/10.7937/TCIA.AXH3-T579} {An expert-annotated dataset of bone marrow cytology in hematologic malignancies}.

\bibitem[{Mina-Osorio(2008)}]{mina2008moonlightingcd13}
Paola Mina-Osorio. 2008.
\newblock The moonlighting enzyme cd13: old and new functions to target.
\newblock \emph{Trends in molecular medicine}, 14(8):361--371.

\bibitem[{Panda et~al.(2021)Panda, Chen, Fan, Sun, Saenko, Oliva, and Feris}]{Panda_Chen_Fan_Sun_Saenko_Oliva_Feris_2021}
Rameswar Panda, Chun-Fu Chen, Quanfu Fan, Ximeng Sun, Kate Saenko, Aude Oliva, and Rogerio Feris. 2021.
\newblock \href {https://doi.org/10.48550/arXiv.2105.05165} {Adamml: Adaptive multi-modal learning for efficient video recognition}.
\newblock (arXiv:2105.05165).
\newblock ArXiv:2105.05165 [cs].

\bibitem[{Pawłowski et~al.(2023)Pawłowski, Wróblewska, and Sysko-Romańczuk}]{Pawłowski_Wróblewska_Sysko-Romańczuk_2023}
Maciej Pawłowski, Anna Wróblewska, and Sylwia Sysko-Romańczuk. 2023.
\newblock \href {https://doi.org/10.3390/s23052381} {Effective techniques for multimodal data fusion: A comparative analysis}.
\newblock \emph{Sensors (Basel, Switzerland)}, 23(5):2381.

\bibitem[{Ronneberger et~al.(2015)Ronneberger, Fischer, and Brox}]{ronneberger2015unet}
Olaf Ronneberger, Philipp Fischer, and Thomas Brox. 2015.
\newblock U-net: Convolutional networks for biomedical image segmentation.
\newblock In \emph{Medical image computing and computer-assisted intervention--MICCAI 2015: 18th international conference, Munich, Germany, October 5-9, 2015, proceedings, part III 18}, pages 234--241. Springer.

\bibitem[{Shaik et~al.(2024)Shaik, Tao, Li, Xie, and Velásquez}]{Shaik_Tao_Li_Xie_Velásquez_2024}
Thanveer Shaik, Xiaohui Tao, Lin Li, Haoran Xie, and Juan~D. Velásquez. 2024.
\newblock \href {https://doi.org/10.1016/j.inffus.2023.102040} {A survey of multimodal information fusion for smart healthcare: Mapping the journey from data to wisdom}.
\newblock \emph{Information Fusion}, 102:102040.

\bibitem[{Stahlschmidt et~al.(2022)Stahlschmidt, Ulfenborg, and Synnergren}]{Stahlschmidt_Ulfenborg_Synnergren_2022}
Sören~Richard Stahlschmidt, Benjamin Ulfenborg, and Jane Synnergren. 2022.
\newblock \href {https://doi.org/10.1093/bib/bbab569} {Multimodal deep learning for biomedical data fusion: a review}.
\newblock \emph{Briefings in Bioinformatics}, 23(2):bbab569.

\bibitem[{Tonge and Caragea(2019)}]{Tonge_Caragea_2019}
Ashwini Tonge and Cornelia Caragea. 2019.
\newblock \href {https://doi.org/10.1145/3308558.3313691} {Dynamic deep multi-modal fusion for image privacy prediction}.
\newblock In \emph{The World Wide Web Conference}, page 1829–1840, San Francisco CA USA. ACM.

\bibitem[{Vaswani et~al.(2017)Vaswani, Shazeer, Parmar, Uszkoreit, Jones, Gomez, Kaiser, and Polosukhin}]{vaswani2017attention}
Ashish Vaswani, Noam Shazeer, Niki Parmar, Jakob Uszkoreit, Llion Jones, Aidan~N Gomez, {\L}ukasz Kaiser, and Illia Polosukhin. 2017.
\newblock Attention is all you need.
\newblock \emph{Advances in neural information processing systems}, 30.

\bibitem[{Wickstr{\"o}m et~al.(2011)Wickstr{\"o}m, Larsson, Nygren, and Gullbo}]{wickstrom2011aminopeptidasecd13}
Malin Wickstr{\"o}m, Rolf Larsson, Peter Nygren, and Joachim Gullbo. 2011.
\newblock Aminopeptidase n (cd13) as a target for cancer chemotherapy.
\newblock \emph{Cancer science}, 102(3):501--508.

\bibitem[{Young et~al.(1999)Young, Steele, Bray, Detmer, Blake, Lucas, and Black~Jr}]{young1999humanc13}
Henry~E Young, Timothy~A Steele, Robert~A Bray, Kristina Detmer, Lisa~W Blake, Paul~W Lucas, and Asa~C Black~Jr. 1999.
\newblock Human pluripotent and progenitor cells display cell surface cluster differentiation markers cd10, cd13, cd56, and mhc class-i.
\newblock \emph{Proceedings of the society for experimental biology and medicine}, 221(1):63--72.

\end{thebibliography}
